\documentclass[journal,twoside,web]{ieeecolor}
\usepackage{generic}
\usepackage{cite}
\usepackage{amsmath,amssymb,amsfonts}
\usepackage{algorithmic}
\usepackage{graphicx}
\usepackage{algorithm,algorithmic}
\usepackage{hyperref}
\hypersetup{hidelinks}
\usepackage{textcomp}
\usepackage{subcaption}
\usepackage{subfloat}
\usepackage{booktabs}
\usepackage{multirow}

\usepackage{bm}
\def\BibTeX{{\rm B\kern-.05em{\sc i\kern-.025em b}\kern-.08em
    T\kern-.1667em\lower.7ex\hbox{E}\kern-.125emX}}
\begin{document}
\title{Dynamic Hypergraph Representation for Bone Metastasis Cancer Analysis}
\author{Yuxuan Chen, Jiawen Li, Huijuan Shi, Yang Xu, Tian Guan, Lianghui Zhu, Yonghong He, Anjia Han
\thanks{This work was supported by National Natural Science Foundation of China (NSFC)(82430062), the Shenzhen Engineering Research Centre (XMHT20230115004), the Shenzhen Science and Technology Innovation Commission (KCXFZ20201221173207022),and the Jilin FuyuanGuan Food Group Co., Ltd. And the internal datasets are collected from The First Affiliated Hostpital of Sun Yat-sen Univeristy. (Yuxuan Chen, Jiawen Li and Huijuan Shi, contributed equally to this work.) (Corresponding authors: Lianghui Zhu; Yonghong He; Anjia Han.)}
\thanks{Yuxuan Chen, Jiawen Li, Tian Guan, Lianghui Zhu and Yonghong He are with the Shenzhen International Graduate School, Tsinghua University, Shenzhen 518055, China. (e-mail: chenyx23@mails.tsinghua.edu.cn; jw-li24@mails.tsinghua.edu.cn; guantian@sz.tsinghua.edu.cn; zhulh@mail.tsinghua.edu.cn; heyh@sz.tsinghua.edu.cn).}
\thanks{Yonghong He is also with the Jinfeng Laboratory, Chongqing, China (e-mail: heyh@sz.tsinghua.edu.cn)}
\thanks{Yang Xu is with the Department of Laboratory Medicine, Shenzhen Children’s Hospital, Shenzhen, 518038, China (e-mail: wdxylioo@gmail.com).}
\thanks{Huijuan Shi, Anjia Han are with the Department of Pathology,
The First Affiliated Hostpital, Sun Yat-sen Univeristy, Guangzhou 510080, China (e-mail: shihj@mail.sysu.edu.cn; hananjia@mail.sysu.edu.cn).}}

\maketitle
\begin{abstract}

Bone metastasis analysis is a significant challenge in pathology and plays a critical role in determining patient quality of life and treatment strategies. The microenvironment and specific tissue structures are essential for pathologists to predict the primary bone cancer origins and primary bone cancer subtyping. By digitizing bone tissue sections into whole slide images (WSIs) and leveraging deep learning to model slide embeddings, this analysis can be enhanced. However, tumor metastasis involves complex multivariate interactions with diverse bone tissue structures, which traditional WSI analysis methods such as multiple instance learning (MIL) fail to capture. Moreover, graph neural networks (GNNs), limited to modeling pairwise relationships, are hard to represent high-order biological associations. To address these challenges, we propose a dynamic hypergraph neural network (DyHG) that overcomes the edge construction limitations of traditional graph representations by connecting multiple nodes via hyperedges. A low-rank strategy is used to reduce the complexity of  parameters in learning hypergraph structures, while a Gumbel-Softmax-based sampling strategy optimizes the patch distribution across hyperedges. An MIL aggregator is then used to derive a graph-level embedding for comprehensive WSI analysis. To evaluate the effectiveness of DyHG, we construct two large-scale datasets for primary bone cancer origins and subtyping classification based on real-world bone metastasis scenarios. Extensive experiments demonstrate that DyHG significantly outperforms state-of-the-art (SOTA) baselines, showcasing its ability to model complex biological interactions and improve the accuracy of bone metastasis analysis.

\end{abstract}


\section{Introduction}
\label{sec:introduction}
\IEEEPARstart{B}{one} metastasis occurs when cancer cells spread from their primary site to bone, leading to significant morbidity and mortality in affected patients \cite{bone1}. The skeletal system is a common target for metastases, especially from cancers such as breast, prostate, lung, and kidney \cite{bone2}. Bone metastases can cause severe pain, pathological fractures, spinal cord compression, and hypercalcemia, severely affecting the quality of life and prognosis of patients \cite{bone3,bone4}.

Bone metastasis cancer analysis is crucial for the administration of precise and effective treatment to patients. Accurate identification of the origin tumor site allows clinicians to tailor therapy specific to the tumor type, thus improving clinical outcomes and potentially extending survival rates \cite{bone5, bone6}. However, the heterogeneity of cancer cells, variations in the microenvironment of metastatic sites, and complex molecular signaling pathways require pathologists to focus on the entire area of the slide, rather than diagnosing based solely on local regions. These challenges contribute to the difficulty of accurately identifying the primary tumor site \cite{bone7}. In addition, bone metastasis cancer often presents with nonspecific clinical symptoms and overlapping imaging features, further complicating the diagnostic process. Despite comprehensive clinical and pathological evaluations, approximately $20\%$ of cases remain difficult to diagnose. As a result, the accurate and rapid analysis of bone metastasis cancer continues to pose a significant challenge in the field of medical science \cite{bone8, bone9, bone10}.

\begin{figure}[t]
\centering
\centerline{\includegraphics[width=1\linewidth]{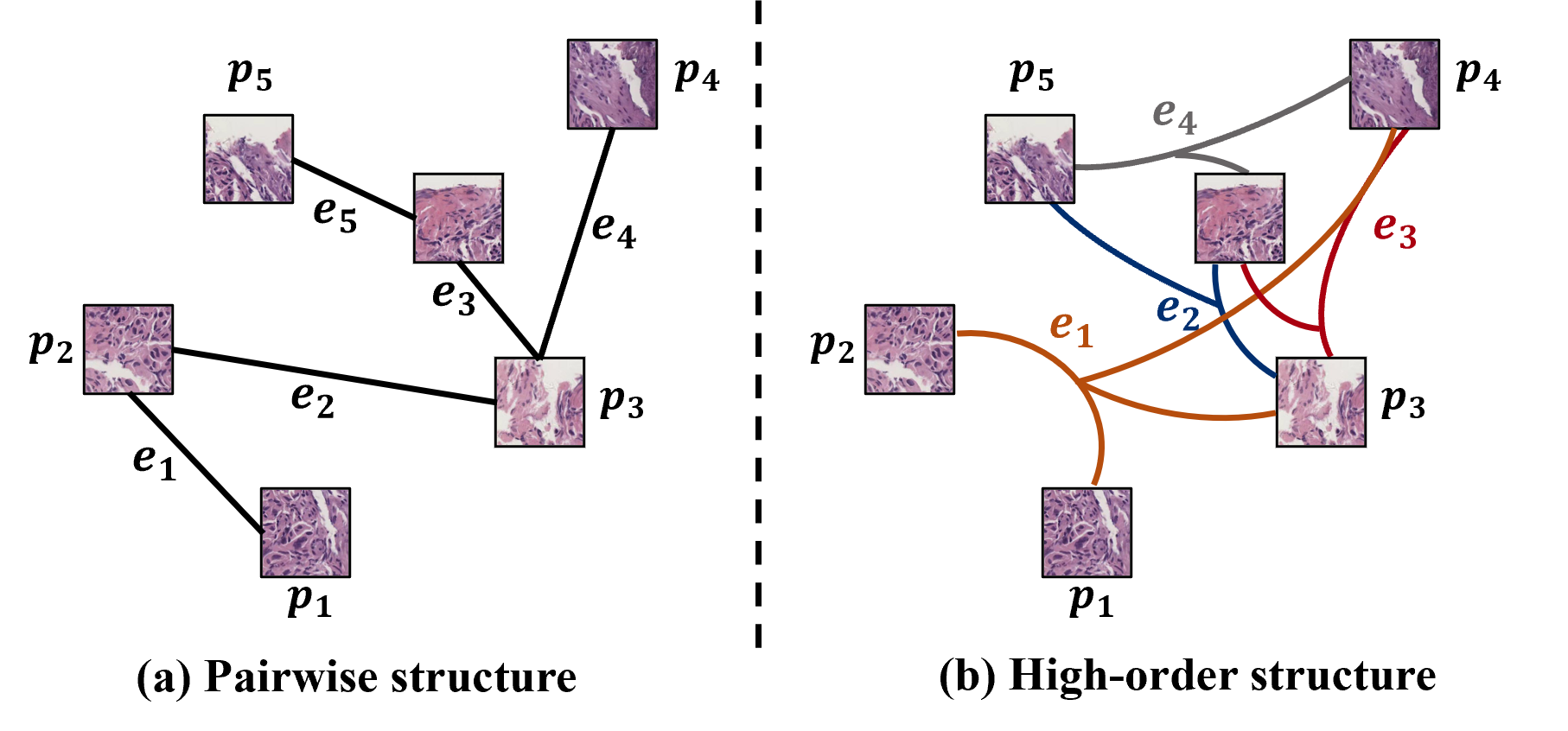}}
\caption{Hypergraph can capture the interactions between distant patches in a more direct way than conventional graph, where $p$ and $e$ denote patch and edge/hyperedge, respectively.}
\label{Intro}
\end{figure}

With the development of optical microscopy scanning systems, physical slides can now be converted to WSI without loss of information, a breakthrough that has been widely recognized by medical institutions and pathologists \cite{wsi1, wsi2}. The digital nature of WSIs has spurred interest in the use of deep learning directly to analyze these images \cite{wsi3, wsi4}. In the context of bone metastasis cancer, it is crucial to analyze the spread of cancer cells within bone tissue. Since cancer cells can spread over various locations in the bone, previous methods \cite{ZLH} focus on annotating regions of interest (ROI) and performing model-based recognition on them. However, such methods require extensive manual annotation, which is both time-consuming and labor-intensive, hindering full automation. Recent advancements in weakly supervised learning, particularly with MIL, demonstrate the potential to learn directly from slide-level labels, achieving success in numerous pathological tasks \cite{ABMIL, CLAM, DSMIL, TransMIL, rrtmil, nciemil, shapley, agent}.

However, in bone metastasis analysis, the high-order multivariate relationships between cells and tissues play a critical role in determining the primary origins and subtyping \cite{bone_1}. Traditional methods relying on simple instance aggregation often fail to capture the intricate interactions between tissue regions, significantly impacting the accuracy of metastasis analysis. GNNs \cite{GNN}, which show excellent performance in capturing complex relationships, offer a promising approach \cite{Patch-GCN, GTP, WiKG}. However, conventional GNNs are limited in that they can only model pairwise relationships through edge construction, without the ability to capture high-order biological associations. Additionally, some graph-based methods, which rely on spatial distance-based relationships, are constrained by the explicit distances between patches, thus limiting their ability to model more complex interactions.

As shown in Fig. \ref{Intro}, conventional pairwise graphs can only model indirect relationships through multiple layers of propagation, such as the path "$p_1 \rightarrow p_2 \rightarrow p_3 \rightarrow p_4$", which may result in the loss of valuable information. Hypergraphs, which allow each hyperedge to connect more than two nodes, offer a more direct way to capture high-order relationships between patches. For example, by learning a hyperedge $e_1$ that connects patches $p_1$, $p_2$, $p_3$, and $p_4$, we can directly capture the interactions between distant patches, as shown in Fig. \ref{Intro}. However, existing hypergraph-based methods face two major limitations: (1) the hypergraph structures they construct using spatial relationships of feature and coordinates are static, meaning they cannot adapt to the evolving knowledge of the model because of lack of learnable parameters; (2) constructing the hypergraph using K-Nearest Neighbors (K-NN) or K-means clustering incurs significant time costs and requires prebuilt hypergraph structures, preventing end-to-end model training.

To address these challenges, we introduce DyHG, a dynamic hypergraph representation tailored for metastasis cancer analysis in bone tissue WSIs. Our approach dynamically constructs the hypergraph incidence matrix using low-rank strategy \cite{low-rank} and Gumbel-Softmax-based sampling\cite{gumbel-softmax}. This allows the model to efficiently explore high-order relationships between patches. The low-rank strategy reduces the number of parameters required to learn the full hypergraph structure by leveraging the initial patch embeddings, while the Gumbel-Softmax-based sampling optimizes the hypergraph structure by enabling efficient exploration of discrete hyperedges. This dynamic construction allows for fine-tuned, end-to-end optimization, which enhances the model's flexibility and ability to capture complex relationships. Furthermore, we use a simple hypergraph convolution network for embedding updates and information aggregation, which consists of node aggregation and hyperedge aggregation. Finally, we perform global graph-level pooling to obtain WSI-level prediction results.

To demonstrate the superior performance of DyHG, we conduct experiments on two large-scale bone metastasis cancer datasets, focusing on two tasks: classifications of primary bone cancer origins and primary bone cancer subtyping. Our experimental results show that DyHG outperforms the SOTA MIL methods. We also perform ablation studies and case analyses that validate the design choices of our model and demonstrate the interpretability of the reference process. In addition, we evaluate the generalizability of DyHG through experiments on two public datasets, where it also outperforms the baseline methods.

In summary, our contributions are as follows:
\noindent
\begin{itemize}
\item[$\bullet$] We introduce DyHG, a dynamic hypergraph representation, for bone metastasis cancer analysis.
\item[$\bullet$] We propose a dynamic hypergraph construction module based on low-rank strategy and Gumbel-Softmax-based sampling to capture high-order relationships between patches.
\item[$\bullet$] Experimental results on two large-scale bone metastasis cancer datasets and two public datasets demonstrate that DyHG outperforms SOTA baselines.
\end{itemize}

\section{Related Work}

\subsection{Embedding-based MIL in the analysis of WSIs}
The paradigm of conventional embedding-based MIL methods involves learning the score for each patch (instance) and aggregating them to obtain a score at the WSI (bag) level for further analysis \cite{MIL1, MIL2}. For example, ABMIL \cite{ABMIL} uses an attention mechanism to learn the attention score of each patch and performs attention-based aggregation to obtain WSI-level scores. CLAM \cite{CLAM} enhances ABMIL by introducing a clustering-constraint loss that improves the model's ability to distinguish between positive and negative instances. DSMIL \cite{DSMIL} introduces a dual-stream architecture with trainable distance measurements to model the relationships between instances, extracting effective representations through contrastive learning \cite{SimCLR}. Inspired by the powerful performance of transformers \cite{transformer}, TransMIL \cite{TransMIL} integrates transformers with MIL to explore both morphological and spatial information. To address the computational overhead and slow inference speed associated with transformer-based MIL methods due to the ultra-long sequences caused by gigapixel-sized images, RetMIL \cite{RetMIL} processes WSI sequences using a hierarchical feature propagation structure. However, all of these embedding-based methods overlook the exploration of interpatch relationships, which is crucial for pathologists in clinical diagnosis.

\subsection{Graph representation learning in digital pathology}
With the rise of GNNs \cite{GCN} and their remarkable achievements in image processing \cite{image1, image2, Image3, image4}, recommendation systems \cite{rec1, rec2, rec3, rec4}, and other fields \cite{che, phy, bio, stock}, several works attempt to apply GNNs in the field of digital pathology to capture the interactive relationships between tissues, cells, and other regions. At the cell level, CGC-Net \cite{CGC-Net} constructs a cell graph based on the spatial locations of cells, coupled with handcrafted features, to model the complex structure of the tissue microenvironment. To explore the relationships between cells and tissues, Hact-Net \cite{hact} proposes a hierarchical cell-to-tissue graph that consists of a low-level cell graph and a high-level tissue graph to capture cell interactions and tissue distribution, respectively. SHGNN \cite{SHGNN} introduces a spatial-hierarchical GNN framework that dynamically constructs both cell and tissue graphs.

In the context of WSI analysis, GNNs treat patches as nodes. For example, Patch-GCN \cite{Patch-GCN} aggregates instance-level histology features hierarchically to model local and global topological structures in the tumor microenvironment using a patch-based graph convolutional network. GTP \cite{GTP} combines graph-based representations with vision transformers to predict disease grade. WiKG \cite{WiKG} represents each WSI as a knowledge graph, dynamically building a directed graph by constructing head and tail nodes for each patch to capture relationships between distant patches. However, these graph-based pathology models are limited to pairwise relationships, making it difficult to capture high-order relationships between patches.

Inspired by the structure of hypergraphs \cite{HGNN, DHNN}, where a hyperedge can connect multiple nodes, several studies focus on applying hypergraphs to digital pathology. For example, b-HGFN \cite{bHGFN} proposes a factorized hypergraph neural network to generate global high-order representations for each WSI. However, this model requires a fixed number of random samples of patches from each WSI, where the number of samples is tied to the model parameters. This constraint means that the number of patches in the input WSI must be at least equal to the number of samples. Hyper-AdaC \cite{Hyper-AdaC} clusters patches based on their features and coordinates, treating the clustered classes as new nodes and using the features and coordinates of patches in each class as the attributes of these new nodes. Hyperedges are then constructed using the K-NN algorithm based on these new node attributes. Updated features are obtained through hypergraph convolution. MaskHGL \cite{MaskHGL} clusters patches based on their features and coordinates to create an embedded hypergraph and a spatial hypergraph, combining the two to obtain a joined hypergraph. The hypergraph learning process is then optimized using a masked hypergraph reconstruction module. However, these hypergraph construction methods, based on K-NN or K-means clustering, require the hypergraph to be constructed in advance and cannot support an end-to-end training process. More importantly, due to the separation between the construction and training processes, the hypergraph structure cannot be dynamically adjusted as the training progresses. The quality of the hypergraph structure is therefore heavily reliant on pre-set hyperparameters.


\begin{figure*}[t]
\centerline{\includegraphics[width=1\linewidth, height=0.4\linewidth]{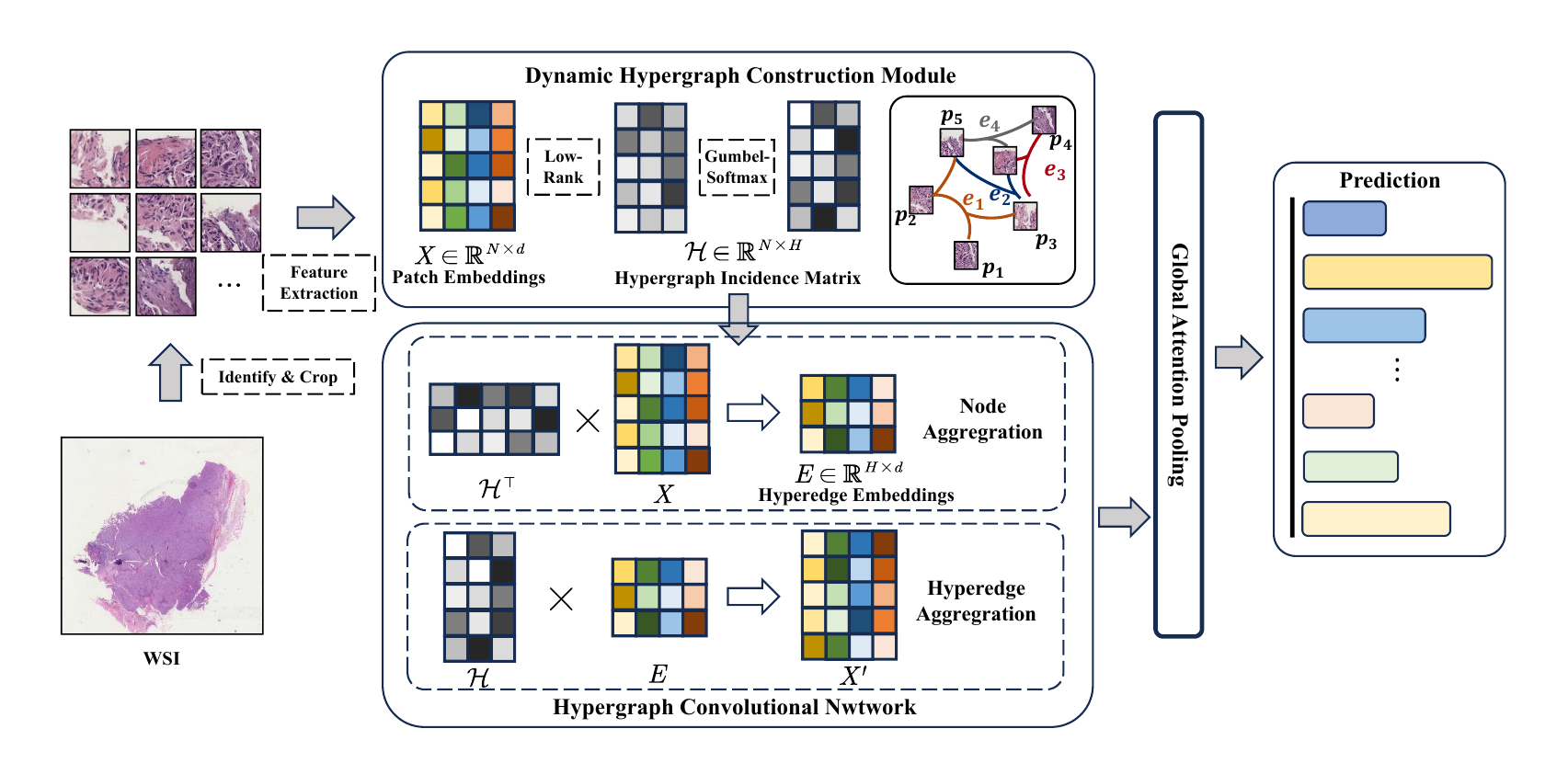}}
\caption{The framework of DyHG. We first preprocess each WSI to obtain the initial embeddings of each patch, then dynamically construct a hypergraph through low-rank and Gumbel-Softmax-based sampling strategy. Next, hypergraph convolutional network is performed to obtain updated patch embeddings. Finally, we obtain the final prediction through global attention pooling.} 
\label{DyHG}
\end{figure*}

\section{Methodology}
In this section, we present DyHG, a dynamic hypergraph representation for bone metastasis cancer analysis, as shown in Fig.\ref{DyHG}. After proprocessing WSIs to obtain initial embeddings, we first introduce a dynamic hypergraph construction module (section \ref{DH}), and then we conduct a hypergraph concolution network on the constructed dynamic hypergraph to propagate information (section \ref{HGCN}). Finally, we show how we obtain the final prediction and the model optimization process (section \ref{pre}).

\subsection{Dynamic hypergraph construction module (DHCM)}
\label{DH}
Given a WSI, we first apply the Otsu thresholding algorithm \cite{otsu} to identify valid tissue areas, followed by a sliding window operation to segment these areas into nonoverlapping patches. A pretrained feature encoder (e.g., UNI \cite{uni}) is then used to extract the initial feature embeddings for each patch. This preprocessing yields an initial patch embedding matrix $\bm{X} = \{\bm{x}_1, \bm{x}_2, \dots, \bm{x}_N\}$ for each WSI, where $N$ denotes the number of patches.

We propose a novel \textbf{D}ynamic \textbf{H}ypergraph \textbf{C}onstruction \textbf{M}odule (DHCM) that addresses the limitations of traditional hypergraph-based methods in pathology. Conventional methods, which often rely on fixed K-NN or K-means clustering, are computationally expensive and produce static hypergraphs that cannot be updated during training. In contrast, DHCM introduces learnable parameters for dynamic hypergraph construction, enabling adaptive learning throughout the training process.

Learning directly the incidence matrix of a hypergraph requires a large number of parameters, specifically $\mathbb{R}^{N\times H}$, where $H$ is the number of hyperedges. To address this challenge, we adopt a low-rank strategy that learns the hypergraph incidence matrix from the initial patch embedding matrix, denoted as:

\begin{equation}
\mathcal{H} = \text{ReLU}(\bm{X} \bm{W}_1),
\label{eq1}
\end{equation}

where $\bm{X} \in \mathbb{R}^{N \times d}$ is the initial patch embedding matrix, $d$ is the dimension of the patch embeddings, and $\bm{W}_1 \in \mathbb{R}^{d \times H}$ is a learnable weight matrix. The resulting $\mathcal{H}_{0} \in \mathbb{R}^{N \times H}$ is the initial hypergraph incidence matrix, where $H$ is the number of hyperedges. This low-rank formulation significantly reduces the number of learnable parameters compared to directly learning a dense incidence matrix, particularly when $d \ll N$.

Although the initial incidence matrix $\mathcal{H}$ provides a starting point for hypergraph construction, it may fail to adaptively focus on critical patches or account for diverse spatial and semantic relationships. To address this limitation, we introduce Gumbel-Softmax for differentiable sampling of hyperedges, allowing the model to dynamically adjust the patch-to-hyperedge assignments during training. Specifically, the soft assignment for each patch is given by:

\begin{equation}
\bm{p}_i = \text{Softmax} \left( \frac{\mathcal{H}_i + \bm{g}_i}{\tau} \right),
\label{eq2}
\end{equation}

where $\mathcal{H}_i \in \mathbb{R}^H$ represents the logits for the $i$-th patch, $\bm{g}_i$ is the Gumbel noise sampled from the Gumbel distribution and $\tau$ is the temperature coefficient. The temperature $\tau$ controls the smoothness of the distribution, allowing the model to balance exploration and exploitation during training. This approach allows each patch to be assigned to multiple hyperedges with continuous probabilities, ensuring a flexible and differentiable learning process.

By leveraging Gumbel-Softmax, DHCM facilitates dynamic hypergraph construction that transcends traditional pairwise relations and static spatial constraints. This enables the model to capture high-order relationships among patches, effectively connecting distant regions that share semantic similarities. In medical pathology, where lesions such as bone metastases may be sparse and dispersed, DHCM excels at associating these regions into a cohesive representation. This dynamic adaptability significantly enhances the model's ability to tackle complex diagnostic tasks, providing a robust foundation for automated clinical decision-making.

The introduction of DHCM represents a substantial advancement in hypergraph-based learning for medical image analysis. Its ability to dynamically adjust the hypergraph structure during training enables the model to better capture nuanced relationships among tissue patches, ultimately improving diagnostic accuracy and generalizability in real-world clinical applications.

\subsection{Hypergraph convolutional network}
\label{HGCN}
Based on the dynamically constructed hypergraph, the next step is to capture the high-level biological correlations between patches through a hypergraph convolutional network. We adopt a simple yet effective hypergraph convolution method, which consists of two main steps: node aggregation and hyperedge aggregation. 

First, based on the incidence matrix constructed in the previous step, we aggregate the features of the nodes contained in each hyperedge to obtain the feature matrix of the hyperedges $\bm{E}\in\mathbb{R}^{H\times d}$, which is defined as:
\begin{equation}
\bm{E} = \text{LeakyReLU}(\mathcal{H}^{\top} \bm{X}),
\label{eq3}
\end{equation}
where $\mathcal{H}\in\mathbb{R}^{N\times H}$ is the incidence matrix, and $\bm{X}\in\mathbb{R}^{N\times d}$ is the node feature matrix. 

Next, through hyperedge aggregation, we aggregate the features of the hyperedges connected to each node to update the node features, obtaining $\bm{X}^{\prime}\in\mathbb{R}^{N\times d}$, which is defined as:
\begin{equation}
\bm{X}^{\prime} = \text{LeakyReLU}(\mathcal{H} \bm{E}).
\label{eq4}
\end{equation}

\subsection{Prediction and optimization}
\label{pre}
After obtaining the initial patch embedding matrix $\bm{X} = \{\bm{x}_{1}, \bm{x}_{2}, \cdots, \bm{x}_{N}\}$ and the updated patch embedding matrix $\bm{X}^{\prime} = \{\bm{x}^{\prime}_{1}, \bm{x}^{\prime}_{2}, \cdots, \bm{x}^{\prime}_{N}\}$, we compute the final embedding for a patch $j$ by averaging its initial and updated embeddings:
\begin{equation}
\bm{x}_j = \frac{1}{2}(\bm{x}_j + \bm{x}^{\prime}_j).
\label{eq5}
\end{equation}
This gives the final patch embedding matrix for each WSI, denoted as $\bm{X} = \{\bm{x}_{1}, \bm{x}_{2}, \cdots, \bm{x}_{N}\}$. To directly predict the primary bone cancer origins and primary bone cancer subtyping, we obtain graph-level embeddings that are used as input to the final classification layer. 

We follow the application of MIL methods in the WSI analysis \cite{CLAM} by performing global attention pooling to compute the attention score for each patch. The graph-level embedding $\bm{h}$ for each WSI is obtained by aggregating the embeddings of all patches weighted by their attention scores:
\begin{equation}
\bm{h} = \sum_{n=1}^N a_n \bm{x}_n,
\label{eq6}
\end{equation}
where $\bm{h} \in \mathbb{R}^{1\times d}$ is the WSI-level embedding, and $a_n$ is the attention score of the $n$-th patch, defined as:
\begin{equation}
a_n = \frac{\text{exp}\{\bm{w} \cdot (\text{tanh}(\bm{V}\bm{x}_n^{\top}) \odot \text{sigmoid}(\bm{U}\bm{x}_n^{\top}))\}}{\sum_{j=1}^N \text{exp}\{\bm{w} \cdot (\text{tanh}(\bm{V}\bm{x}_j^{\top}) \odot \text{sigmoid}(\bm{U}\bm{x}_j^{\top}))\}},
\label{eq7}
\end{equation}
where $\bm{w} \in \mathbb{R}^{1 \times M}$, $\bm{V}, \bm{U} \in \mathbb{R}^{M \times d}$ are learnable weight matrices, $M$ is the hidden dimension, and we set $M=256$.

Finally, the classification is performed by mapping the graph-level embedding $\bm{h}$ through a fully connected layer followed by a softmax function to obtain the probability scores:
\begin{equation}
\bm{\hat{y}} = \text{Softmax}(\bm{h}\bm{W}),
\label{eq8}
\end{equation}
where $\bm{\hat{y}} \in \mathbb{R}^{1\times C}$ is the prediction result for each WSI, $\bm{W} \in \mathbb{R}^{d \times C}$ is a learnable weight matrix, and $C$ is the number of categories. 

To optimize the model, we employ cross-entropy loss \cite{celoss}, defined as:
\begin{equation}
\mathcal{L} = -\frac{1}{P} \sum_{p=1}^{P} \sum_{c=1}^{C} y_{p,c} \ln \hat{y}_{p,c},
\label{eq9}
\end{equation}
where $\mathcal{L}$ is the loss, $P$ is the number of samples, $y_{p,c}$ is the one-hot ground truth label, and $\hat{y}_{p,c}$ is the predicted probability.

\begin{figure}[t]
\centering
\centerline{\includegraphics[width=0.8\linewidth]{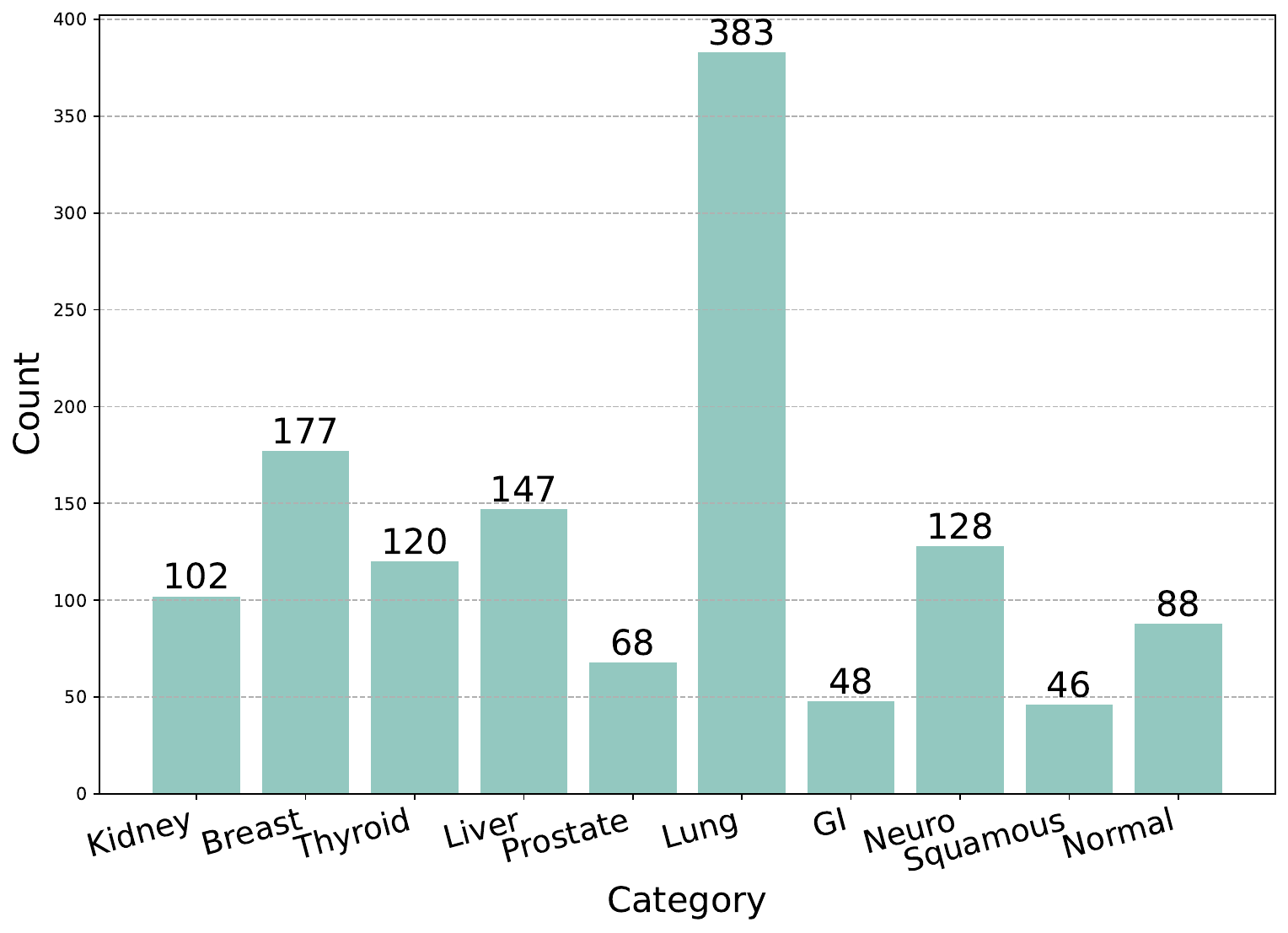}}
\caption{Data statistics of bone metastasis cancer data.}
\label{data}
\end{figure}

\begin{figure}[t]
\centering
\centerline{\includegraphics[width=0.8\linewidth]{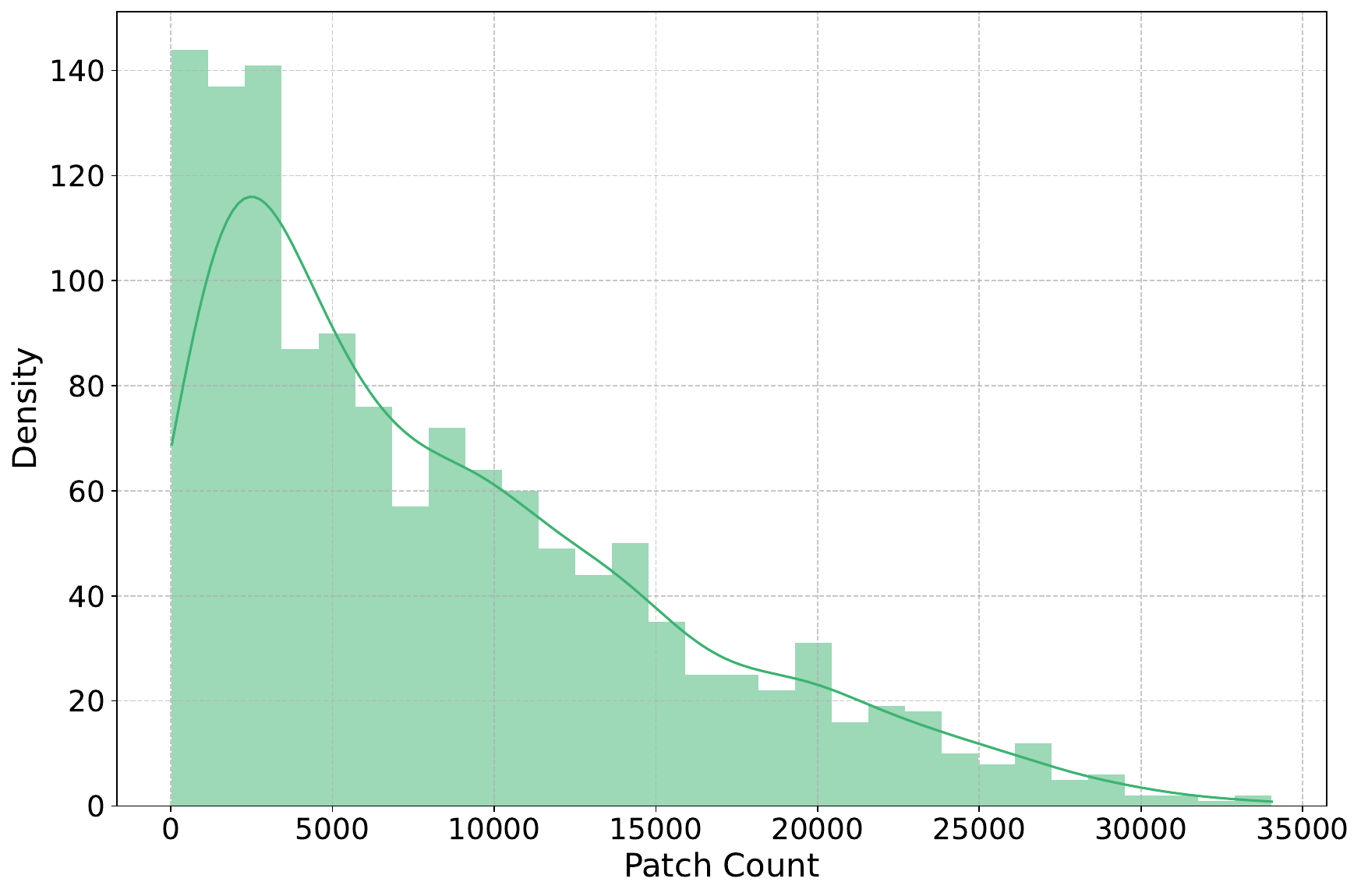}}
\caption{Distribution of patch counts per WSI.}
\label{distribution}
\end{figure}

\section{Experiments}
\begin{table*}[ht]
    \centering
    \caption{Comparison results on different classification tasks. Bold and underline indicate the best and second best performance among all models, respectively. And the number in the bottom right corner represents the standard deviation obtained from five repeated experiments with different random seeds. 'Acc' stands for 'accuracy' and 'Bal acc' stands for 'balanced accuracy'.}
    \label{tab1}
    \begin{tabular}{lcccccccccccc}
        \toprule
        \multirow{2}{*}{\textbf{Method}} & \multicolumn{4}{c}{\textbf{Primary Bone Cancer Origins}} & & \multicolumn{4}{c}{\textbf{Primary Bone Cancer subtyping}} \\
        \cmidrule{2-5} \cmidrule{7-10}
        & \textbf{Acc} & \textbf{Bal-acc} & \textbf{Specificity} & \textbf{Weighted F1} & & \textbf{Acc} & \textbf{Bal-acc} & \textbf{Specificity} & \textbf{Weighted F1}\\
        \midrule
        ABMIL \cite{ABMIL} & $75.77_{1.81}$ & $68.25_{2.70}$ & $96.26_{0.25}$ & $74.51_{1.96}$ & & $86.85_{2.70}$ & $84.22_{5.90}$ & $92.52_{0.87}$ & $89.13_{3.24}$\\
        CLAMSB \cite{CLAM} & $84.00_{1.06}$ & $79.25_{1.36}$ & $97.54_{0.14}$ & $83.87_{1.13}$ & & $\underline{93.82_{0.59}}$ & $81.97_{1.09}$ & $\underline{95.64_{0.26}}$ & $\underline{93.75_{0.60}}$\\
        CLAMMB \cite{CLAM} & $\underline{85.04_{1.01}}$ & $\underline{80.38_{0.43}}$ & $\underline{97.70_{0.11}}$ & $\underline{84.88_{0.94}}$ & & $92.35_{1.22}$ & $82.98_{3.80}$ & $95.43_{0.42}$ & $92.45_{0.99}$ \\
        TransMIL \cite{TransMIL} & $83.13_{1.51}$ & $77.32_{1.14}$ & $97.35_{0.19}$ & $82.91_{1.38}$ & & $91.09_{2.63}$ & $75.29_{3.91}$ & $92.89_{1.05}$ & $90.89_{2.13} $ \\
        DSMIL \cite{DSMIL} & $82.96_{1.35}$ & $77.74_{1.10}$ & $97.33_{0.21}$ & $82.84_{1.25} $ & & $93.11_{1.04}$ & $79.63_{3.48}$ & $94.25_{0.93}$ & $92.90_{1.08}$ \\
        RRTMIL \cite{rrtmil} & $83.30_{1.08}$ & $77.85_{2.06}$ & $97.36_{0.23}$ & $83.05_{1.16} $ & & $92.46_{1.33}$ & $84.26_{3.81}$ & $94.61_{0.99}$ & $92.46_{1.28}$ \\
        WiKG \cite{WiKG} & $79.88_{1.43}$ & $74.75_{0.95}$ & $96.91_{0.17}$ & $79.85_{1.27} $ & & $90.43_{1.71}$ & $78.32_{2.20}$ & $93.09_{0.91}$ & $90.35_{1.45}$ \\
        PatchGCN \cite{Patch-GCN} & $80.00_{2.23}$ & $74.68_{1.26}$ & $96.92_{0.27}$ & $79.93_{2.00}$ & & $85.37_{3.61}$ & $82.91_{1.68}$ & $93.97_{0.81}$ & $86.88_{2.90}$ \\
        Hyper-AdaC \cite{Hyper-AdaC} & $83.54_{1.91}$ & $78.23_{1.36}$ & $97.42_{0.30}$ & $83.38_{1.72}$ & & $92.35_{1.52}$ & $\underline{85.50_{1.52}}$ & $95.29_{0.38}$ & $92.49_{1.29}$ \\
        bHGFN \cite{bHGFN} & $73.61_{4.27}$ & $66.86_{3.37}$ & $95.80_{0.37}$ & $72.15_{4.16}$ & & $87.52_{1.86}$ & $66.97_{5.32}$ & $91.57_{1.21}$ & $86.28_{2.12}$ \\
        \midrule
        \textbf{DyHG(Ours)} & $\mathbf{86.32_{0.96}}$ & $\mathbf{81.02_{0.56}}$ & $\mathbf{97.86_{0.18}}$ & $\mathbf{86.13_{0.92}}$ & & $\mathbf{94.08_{0.44}}$ & $\mathbf{86.06_{2.02}}$ & $\mathbf{96.15_{0.42}}$ & $\mathbf{94.11_{0.43}}$ \\
        \bottomrule
    \end{tabular}
\end{table*}

\subsection{Datasets}
To verify the ability of our proposed DyHG in the bone metastasis analysis task, we collect large-scale bone metastasis cancer data from the First Affiliated Hospital of Sun Yat-sen University, and the data statistics are shown in Fig. \ref{data}. According to clinical diagnosis criteria, we divide this batch of data into two datasets, one is the primary bone cancer origins dataset, including eight categories: kidney, breast, thyroid, liver, prostate, lung, gastrointestinal cancer (GI), as well as normal tissue sections, and the other is the primary bone cancer subtyping dataset, including four categories: adenocarcinoma (AdCa), neuroendocrine (Neuro), squamous and Normal. Specifically, except normal. category, all other categories in the primary tumor type dataset belong to AdCa. These data are scanned using a Teksqray scanner to form sdpc-format files, and then the open-source sdpc-python package is used for patch preprocessing. The average number of patches is 8696, and the specific number of patches is represented by a histogram, as shown in Fig. \ref{distribution}. It can be seen that the number of patches contained in the WSI of our dataset ranges from dozens to more than 30,000, which can comprehensively measure the performance of the model in processing WSIs of different sizes.

\subsection{Baselines}
We compare DyHG with several SOTA baselines including embedding-based MIL and graph representation learning MIL, the brief introduction of these baselines is as follows:

\noindent \textbf{(1) Embedding-based MIL baselines}.
\begin{itemize}
\item[$\bullet$] \textbf{ABMIL} \cite{ABMIL}: An MIL framework that aggregates patch embeddings based on attention score.
\item[$\bullet$] \textbf{CLAM} \cite{CLAM}: It further optimizes ABMIL by introducing instance clustering loss, where CLAMSB uses a single attention branch to focus on one dominant instance per slide, while CLAMMB employs multiple attention branches to capture diverse ROIs within the slide.
\item[$\bullet$] \textbf{TransMIL} \cite{TransMIL}:A transformer-based MIL framework that treats patches as sequence. 
\item[$\bullet$] \textbf{DSMIL} \cite{DSMIL}: It proposes a dual-stream architecture to capture the relations between instances.
\item[$\bullet$] \textbf{RRTMIL} \cite{rrtmil}: AB-MIL is further optimized by reinforcing the instance features online.
\end{itemize}
\noindent \textbf{(2) Graph representation learning baselines}.
\begin{itemize}
\item[$\bullet$] \textbf{WiKG} \cite{WiKG}:A dynamic directed graph construction based MIL framework that treats WSI as a knowledge graph. 
\item[$\bullet$] \textbf{Patch-GCN} \cite{Patch-GCN}: It treats patches as nodes to perform patch-based graph convolutional network.
\item[$\bullet$] \textbf{Hyper-AdaC} \cite{Hyper-AdaC}: An adaptive clustering-based hypergraph representation to model high-order correlations among different regions of the WSIs.
\item[$\bullet$] \textbf{bHGFN} \cite{bHGFN}: a factorization neural network that embeds the correlation among large-scale vertices and hyperedges in two
low-dimensional latent semantic spaces separately, empowering dense sampling.
\end{itemize}

\subsection{Implementation Details}
For all WSIs, we crop each of them into $256\times256$ patches at 20x magnification, then UNI\cite{uni} is used as the frozen feature extractor to extract the initial features of each patch with a dimension of 1024. For each category of the two bone metastasis datasets, we divide the data into training set, validation set, and test set in a ratio of 5:2:3. All of these models are trained on an Nvidia RTX A6000 GPU with Pytorch library. We set the batch size as 1, the number of epochs as 50, and Adam optimizer \cite{Adam} is used with a learning rate of $10^{-4}$ and a weight decay of $10^{-5}$. For DyHG, the only hyperparameters we need to adjust are the number of hyperedges $H$ and the temperature coefficient $\tau$. For the primary bone cancer origin classification task, we set $H$ as $20$ and $\tau$ as $0.1$, respectively. For the primary bone cancer subtyping classification task, we set $H$ as $16$ and $\tau$ as $0.15$, respectively. For all baselines, we perform experiments in the same settings. 

\subsection{Evaluation Metrics}
To evaluate the performance of DyHG and baseline methods, we use four metrics: \textbf{accuracy}, \textbf{balanced accuracy}, \textbf{specificity}, and \textbf{Weighted F1}. These metrics comprehensively assess overall performance, the ability to handle imbalanced data, and the precision recall trade-off. During the training process, the model that achieves the best balanced accuracy in the validation set is selected for evaluation in the test set.

\section{Results}

\subsection{Comparison Results on Two Tasks}
We compare DyHG with several baselines on primary bone cancer origins and primary bone cancer subtyping classification tasks. The results are shown in Table \ref{tab1}. The following observations can be made:

(1) \textbf{Superior overall performance:} DyHG achieves the highest scores in all four metrics in both tasks, leading the best performing baselines by 0.16\% to 1.28\%. For example, in the primary bone cancer origins classification task, DyHG outperforms CLAMMB by 0.64\% in accuracy and 0.64\% in balanced accuracy. In the primary bone cancer subtyping classification task, DyHG achieves 94.08\% accuracy and 86.06\% balanced accuracy, surpassing CLAMSB by 0.26\% and 4.09\%, respectively. These results highlight the effectiveness of the dynamic hypergraph construction module, which better captures high-order biological correlations in pathological images.

(2) \textbf{Improved stability:} DyHG demonstrates lower standard deviations compared to other methods in 7 out of 8 metrics in both tasks. For example, in the primary bone cancer origins classification task, DyHG achieves a std of only 0.44 in accuracy, significantly lower than ABMIL (2.70), CLAMMB (1.22) and TransMIL (2.63). This stability ensures consistent performance across random seeds, which is critical for clinical applications such as bone metastasis diagnosis, where reliability is essential.

(3) \textbf{Comparison with transformer-based MIL baseline:} Transformer can also capture the interaction between patches. Compared to TransMIL, a transformer-based MIL model, DyHG achieves significantly better results, particularly in balancer accuracy (a 3.7\% improvement in the primary bone cancer origins classification task and a 10.8\% improvement in the primary bone cancer subtyping task). This demonstrates the limitation of treating patches as sequential tokens in TransMIL, which struggles to capture spatial high-order relationships effectively. In contrast, DyHG’s dynamic hypergraph construction explicitly captures high-order correlations and heterogeneous spatial patterns, making it better suited for tasks involving complex tumor patterns, such as bone metastasis classification.

(4) \textbf{Comparison with graph representation learning baselines:} Among graph-based methods, Hyper-AdaC achieves relatively strong performance, confirming the utility of hypergraph representations in computational pathology. DyHG further enhances performance by dynamically constructing hypergraphs that adapt to the heterogeneity of WSIs. For example, DyHG achieves the highest specificity (97.86\%) in the primary bone cancer origins classification task, surpassing Hyper-AdaC (97.42\%). This highlights the advantage of DyHG's sampling strategy, which ensures more informative hyperedges. On the other hand, bHGFN performs poorly, probably due to its reliance on random patch sampling, which loses critical biological information.


\begin{figure}[ht]
\centering
\centerline{\includegraphics[width=1\linewidth]{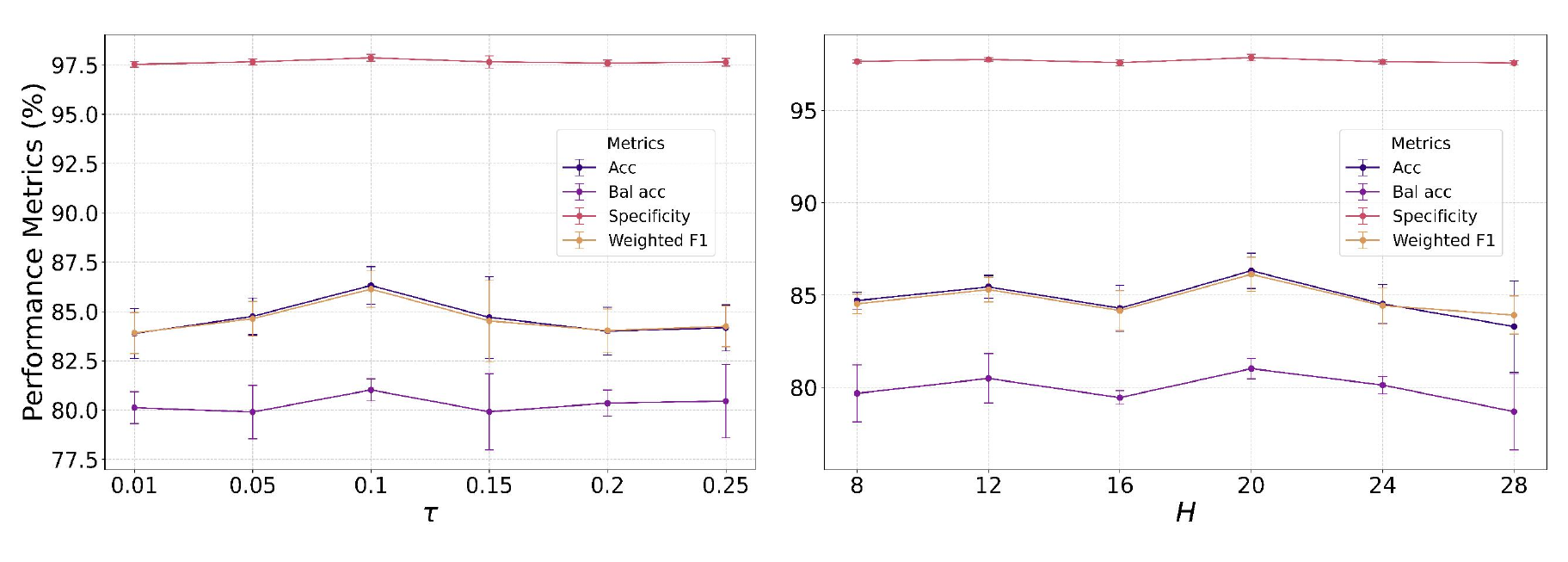}}
\caption{Hyperparameters analysis on primary bone cancer origins classification task}
\label{ha_type}
\end{figure}

\begin{figure}[ht]
\centering
\centerline{\includegraphics[width=1\linewidth]{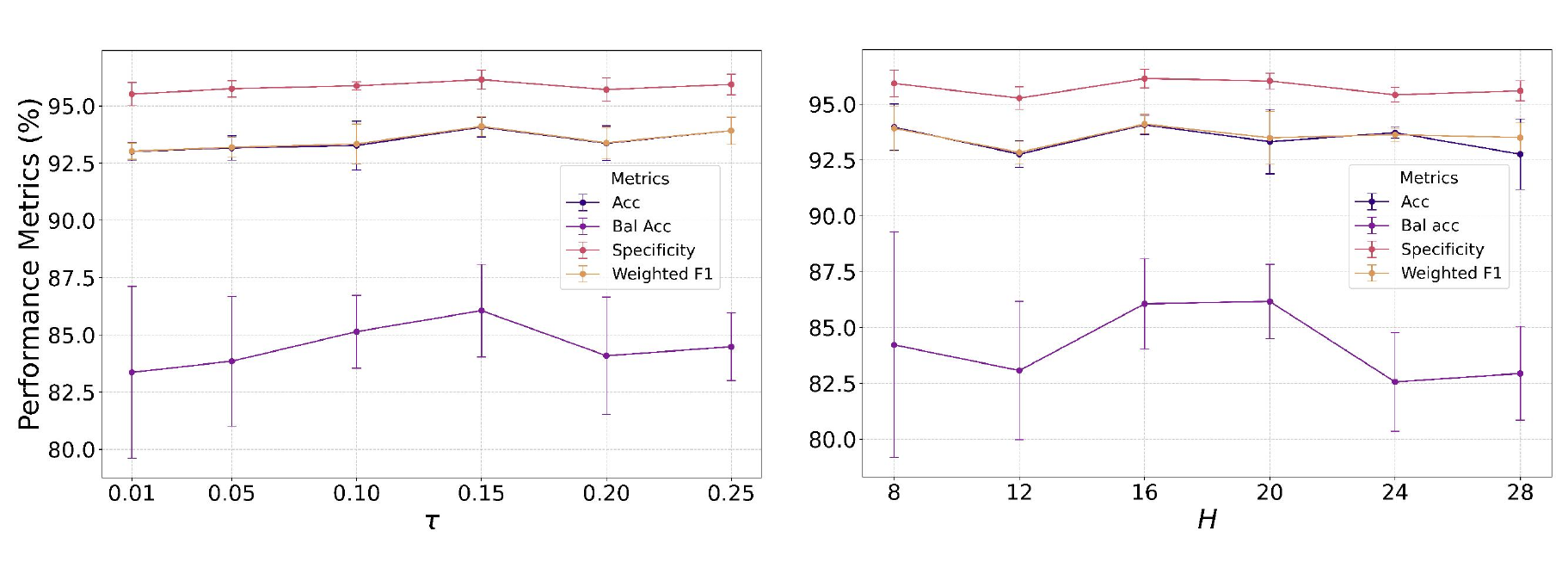}}
\caption{Hyperparameters analysis on primary bone cancer subtyping classification task}
\label{ha_morph}
\end{figure}

\subsection{Hyperparameter Analysis}
DyHG involves two key hyperparameters that require adjustment: the number of hyperedges ($H$) and the temperature coefficient ($\tau$). These hyperparameters directly influence the learning quality of the hypergraph structure by controlling two aspects: structure size and patch distribution. For the number of hyperedges $H$, we vary it within the range $\{8, 12, 16, 20, 24, 28\}$, and for the temperature coefficient $\tau$, we vary it within the range $\{0.01, 0.05, 0.1, 0.15, 0.2, 0.25\}$. The experimental results are illustrated in Fig.\ref{ha_type} and Fig.\ref{ha_morph}. The following observations can be made about the results:  

(1) \textbf{Impact of the number of hyperedges ($H$):} As $H$ increases, the performance of DyHG first improves and then declines, suggesting an optimal value within the tested range. When $H$ is too small, each hyperedge contains an excessive number of patches, resulting in overly broad relationships within the hyperedge. This broadness prevents the model from capturing the subtle distinctions between different regions in pathological images, such as the fine-grained differences between the lesion and normal tissues, ultimately reducing the model's discriminative power. In contrast, when $H$ is too large, the information within each hyperedge becomes too sparse, leading to an overly fragmented hypergraph structure. This fragmentation hinders the model's ability to capture the global features of key lesions, thereby diminishing recognition accuracy.

(2) \textbf{Impact of the temperature coefficient ($\tau$):} The model's performance exhibits a similar trend with $\tau$ as with $H$: an initial increase followed by a decline. When $\tau$ is too low, the Gumbel-Softmax-based sampling causes the patch weights to be disproportionately concentrated on a single hyperedge. This overconcentration makes the model overly reliant on a few hyperedges while neglecting other potentially relevant ones, leading to overfitting and an inability to capture the complex spatial relationships inherent in pathological images. On the other hand, when $\tau$ is too high, the patch weights are distributed too evenly across multiple hyperedges, resulting in a lack of specificity. This uniformity impairs the model's ability to effectively capture the local features of key lesion areas, reducing its capacity to distinguish between pathological regions, and ultimately degrading overall performance.

\begin{figure}[t]
\centering
\centerline{\includegraphics[width=0.8\linewidth]{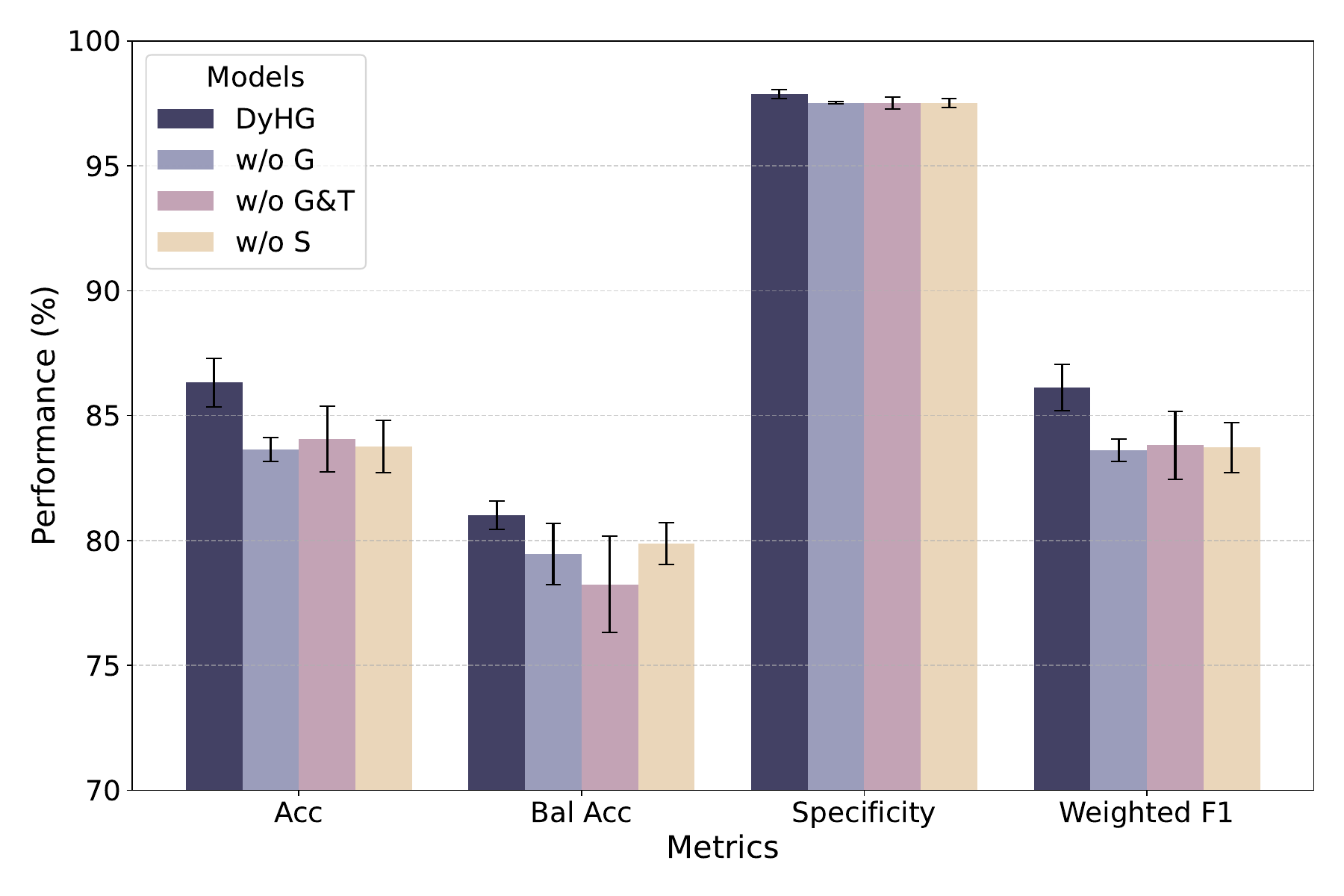}}
\caption{Ablation study on primary bone cancer origins classification task.}
\label{t_as}
\end{figure}

\begin{figure}[t]
\centering
\centerline{\includegraphics[width=0.8\linewidth]{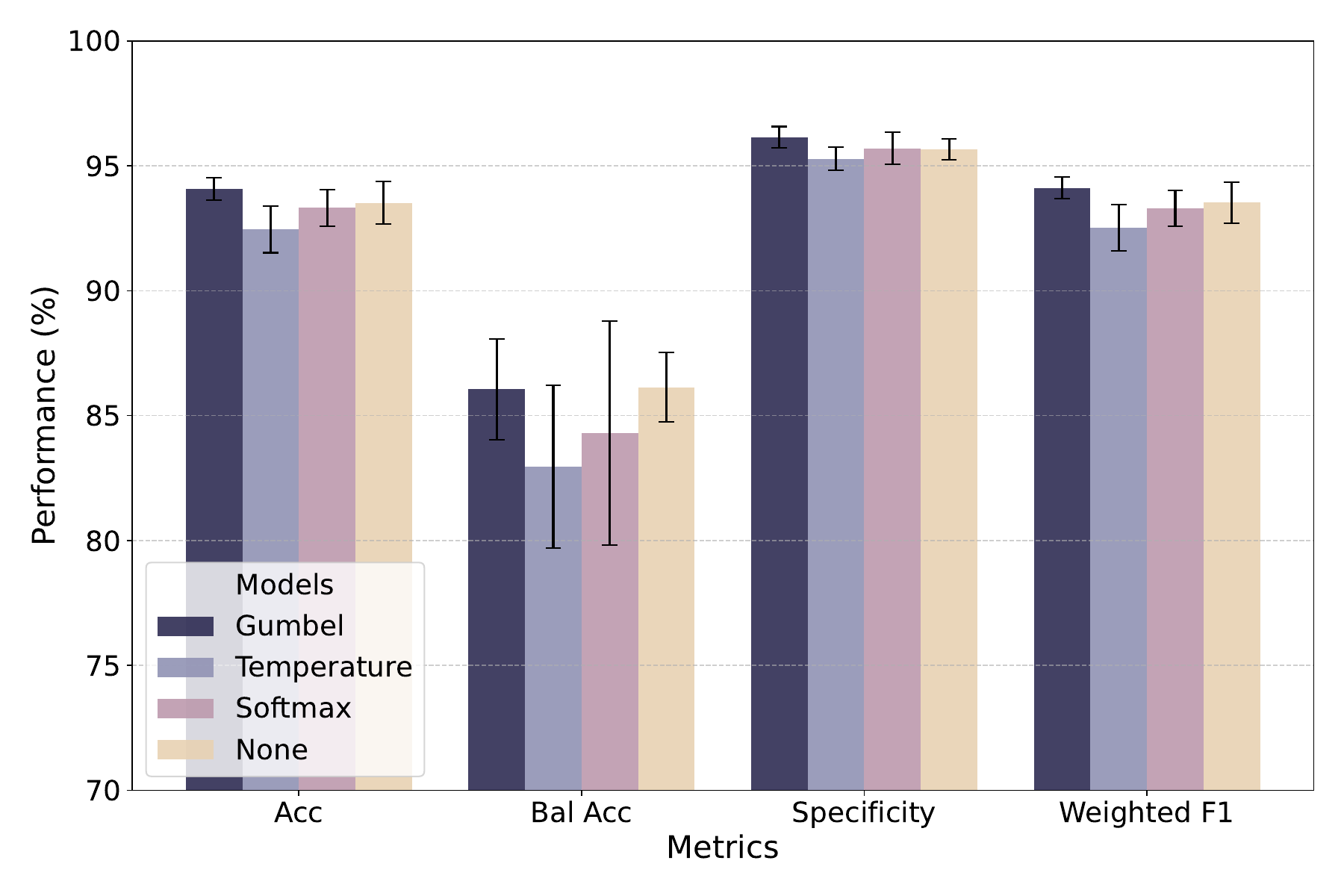}}
\caption{Ablation study on primary bone cancer subtyping classification task.}
\label{m_as}
\end{figure}

\subsection{Effectiveness of the Gumbel-Softmax-based sampling}
To evaluate the effectiveness of Gumbel-Softmax-based sampling, we designed three additional variants of our model:
\begin{itemize}
    \item \textbf{w/o G:} The noise from the Gumbel distribution is removed, converting the sampling process into a deterministic softmax sampling based on the temperature coefficient (Eq. \ref{eq2}).
    \item \textbf{w/o G\&T:} Ordinary softmax is used directly for sampling and optimization, without incorporating temperature adjustment or Gumbel noise.
    \item \textbf{w/o S:} The hypergraph structure learned by the low-rank strategy is directly used for subsequent learning without any sampling optimization.
\end{itemize}
The results of the ablation study for the origin tumor type classification task and the origin tumor morphology classification task are shown in Fig.~\ref{t_as} and Fig.~\ref{m_as}, respectively. 

The following observations can be made from the results:
(1) \textbf{DyHG achieves superior performance:} In both tasks, DyHG consistently outperforms the three variants. This demonstrates the effectiveness of Gumbel noise in improving hypergraph-based modeling, as it introduces controlled randomness during sampling, which allows the model to better explore potential hyperedge configurations and avoid overfitting to local optima. Furthermore, the learned hyperedge assignments align more closely with the complex spatial relationships in WSIs of bone metastases, particularly the heterogeneous distribution of tumor and non-tumor patches.

(2) \textbf{Performance degradation without Gumbel-Softmax sampling:} Both the w/o G and w/o G\&T variants exhibit significant performance drops across all metrics. Without Gumbel noise, softmax sampling becomes entirely deterministic, limiting its ability to explore alternative hyperedge configurations. Moreover, this deterministic behavior leads to homogeneous hyperedges, which fail to capture the intricate patch-level variations and biological associations critical for distinguishing subtle differences between lesion and normal tissue. 


(3) \textbf{Surprising performance of the w/o S variant:} Interestingly, the w/o S variant, which skips the sampling entirely, performs better than both the w/o G and w/o G\&T variants. This suggests that the hypergraph structure learned through the low-rank strategy already provides a strong foundation for modeling high-order relationships between patches, while inappropriate sampling may have side effects. However, its performance still lags behind DyHG, indicating that the incorporation of Gumbel-Softmax sampling further enhances the model’s ability to capture complex biological interactions and patch-level diversity, leading to a more accurate and robust representation.

\begin{figure}[t]
\centering
\centerline{\includegraphics[width=1\linewidth]{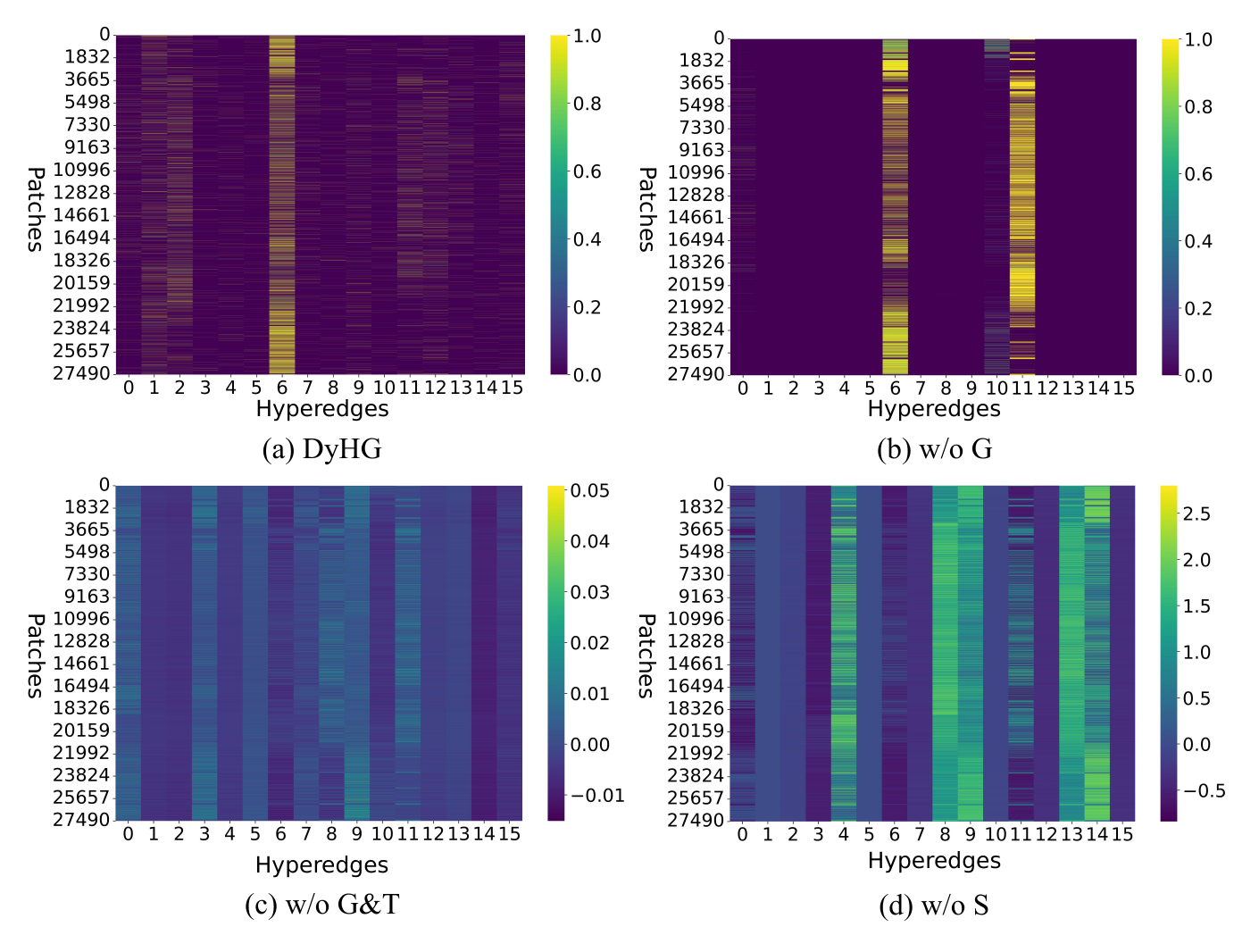}}
\caption{heatmaps of incidence matrices. Where (a) corresponds to DyHG, (b) w/o G, (c) w/o G\&T, and (d) w/o S}
\label{i_m}
\end{figure}

We further visualize the heatmaps of the incidence matrices constructed by four different methods for the same WSI in the test set, reflecting the distribution of patches across hyperedges. The results are shown in Fig.~\ref{i_m}. These visualizations provide valuable insights into the effects of different sampling strategies on hyperedge construction in the context of bone metastasis, where it is critical to capture subtle interactions between pathological regions.

(1) For w/o G (Fig.~\ref{i_m}(b)), where the noise from the Gumbel distribution is removed, the sampling process becomes deterministic, reducing the diversity of patch distributions across hyperedges. This limits the model's ability to capture the spatial heterogeneity required for identifying tumor regions in bone metastasis. (2) For w/o G\&T (Fig.~\ref{i_m}(c)), where only ordinary softmax is used, the heatmap shows nearly uniform patch weights across the hyperedges. This uniformity fails to capture the biological heterogeneity in pathological images, making it harder to distinguish subtle tumor patterns. (3) For w/o S (Fig.~\ref{i_m}(d)), where the hypergraph structure is used without additional optimization, the heatmap reveals multiple hyperedges with invalid weights exceeding 1 or less than 0. This results in poorly defined hyperedges, reducing the model’s ability to prioritize biologically meaningful patches. (4) In contrast, DyHG (Fig.~\ref{i_m}(a)) achieves a balanced and diverse distribution of patch weights across hyperedges, thanks to the Gumbel-Softmax-based sampling. This enables DyHG to effectively model complex spatial relationships, making it better suited for capturing heterogeneous tumor regions in bone metastasis.

\subsection{Time efficiency of the proposed DHCM}
\begin{figure}[t]
\centering
\centerline{\includegraphics[width=0.8\linewidth]{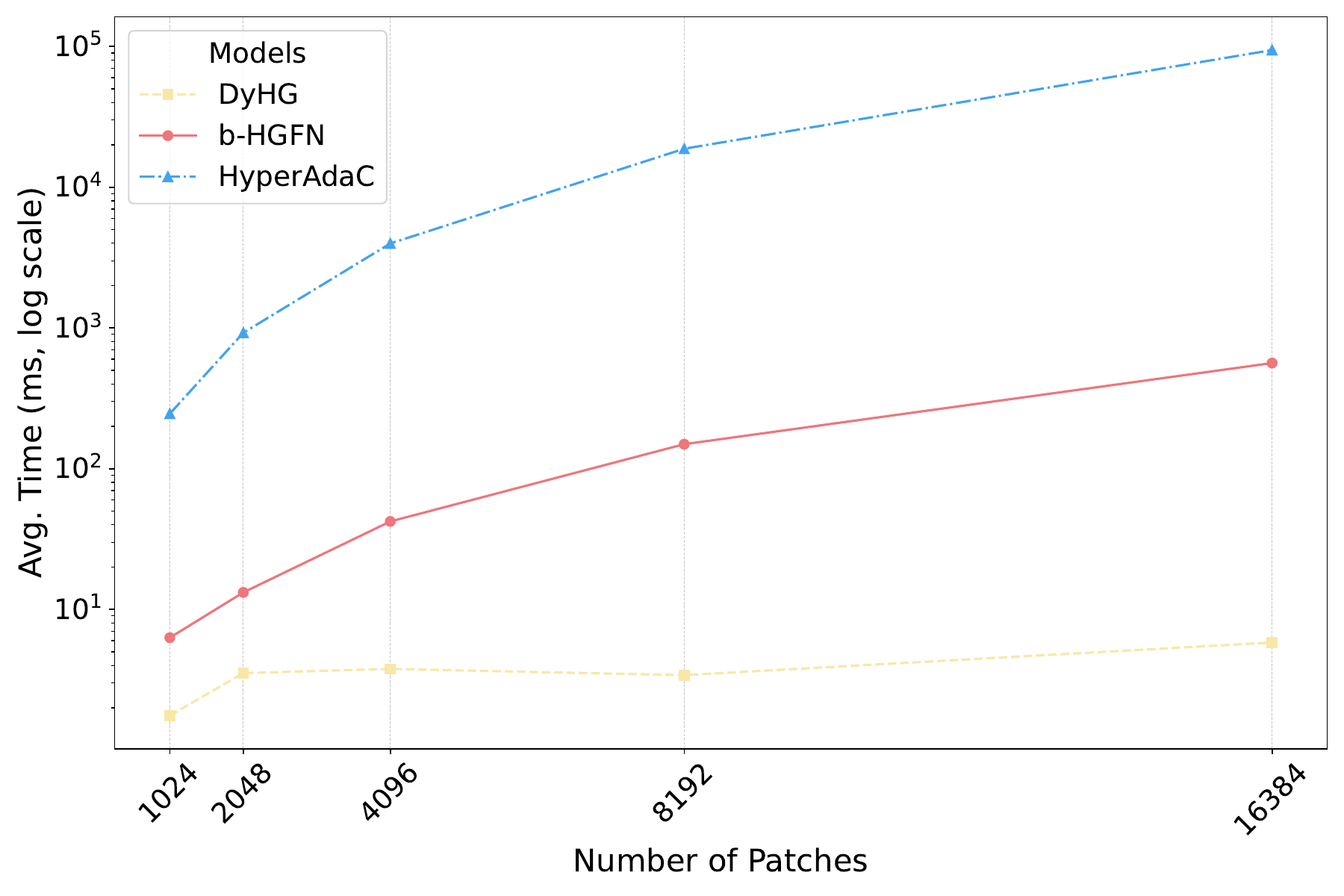}}
\caption{Comparison of the time required to construct a hypergraph of WSI containing different numbers of patches}
\label{t_compare}
\end{figure}

To assess the time efficiency of the proposed DHCM, we compare its hypergraph construction time with that of Hyper-AdaC and bHGFN on WSIs containing varying numbers of patches under identical conditions. The results are presented in Fig.~\ref{t_compare}. As shown in the figure, the hypergraph construction time for DHCM remains relatively stable as the number of patches increases. In contrast, bHGFN and Hyper-AdaC, which rely on computationally expensive methods such as k-NN and K-means clustering for hypergraph construction, exhibit exponential growth in time consumption with increasing patch numbers. This makes these methods less practical for handling large-scale WSIs, particularly in clinical settings where efficiency is critical. 

These results highlight the superior time efficiency of the proposed DHCM, demonstrating its scalability and practicality for real-world clinical applications. By maintaining consistent computational efficiency across varying WSI sizes, DHCM holds significant promise for deployment in clinical workflows where rapid and reliable processing of large pathological images is essential.

\begin{table*}[ht]
    \centering
    \caption{Comparison results on two public datasets. Bold and underline indicate the best and second best performance among all models, respectively. And the number in the bottom right corner represents the standard deviation obtained from five repeated experiments with different random seeds. 'Acc' stands for 'accuracy' and 'Bal acc' stands for 'balanced accuracy'.}
    \label{tab2}
    \begin{tabular}{lcccccccccccc}
        \toprule
        \multirow{2}{*}{\textbf{Method}} & \multicolumn{4}{c}{\textbf{CAMELYON+}} & & \multicolumn{4}{c}{\textbf{PANDA}} \\
        \cmidrule{2-5} \cmidrule{7-10}
        & \textbf{Acc} & \textbf{Bal-acc} & \textbf{Specificity} & \textbf{Weighted F1} & & \textbf{Acc} & \textbf{Bal-acc} & \textbf{Specificity} & \textbf{Weighted F1}\\
        \midrule
        ABMIL \cite{ABMIL} & $87.27_{0.57}$ & $63.01_{1.08}$ & $94.64_{0.17}$ & $85.49_{0.56}$ & & $63.50_{0.74}$ & $58.52_{0.57}$ & $92.64_{0.12}$ & $63.28_{0.83}$\\
        CLAMSB \cite{CLAM} & $85.45_{0.77}$ & $63.44_{1.90}$ & $94.40_{0.25}$ & $83.75_{1.48}$ & & $64.84_{0.62}$ & $60.29_{0.42}$ & $92.86_{0.11}$ & $64.55_{0.65}$\\
        CLAMMB \cite{CLAM} & $87.13_{0.80}$ & $66.26_{1.45}$ & $94.58_{0.24}$ & $86.27_{0.72}$ & & $65.83_{0.53}$ & $61.60_{0.58}$ & $93.06_{0.11}$ & $65.48_{0.48}$ \\
        TransMIL \cite{TransMIL} & $\underline{87.62_{1.68}}$ & $66.00_{0.84}$ & $94.74_{0.25}$ & $\underline{86.50_{0.89}}$ & & $59.15_{1.12}$ & $53.21_{0.75}$ & $91.71_{0.12}$ & $58.33_{0.53} $ \\
        DSMIL \cite{DSMIL} & $82.90_{3.14}$ & $61.23_{2.38}$ & $92.67_{0.91}$ & $82.60_{1.90} $ & & $62.90_{0.40}$ & $57.97_{0.69}$ & $92.49_{0.09}$ & $62.55_{0.58}$ \\
        RRTMIL \cite{rrtmil} & $82.65_{1.68}$ & $61.89_{2.47}$ & $93.43_{0.67}$ & $82.57_{1.66} $ & & $64.36_{0.80}$ & $59.86_{0.49}$ & $92.79_{0.13}$ & $64.14_{0.71}$ \\
        WiKG \cite{WiKG} & $83.19_{3.70}$ & $66.81_{3.74}$ & $93.75_{1.13}$ & $84.00_{2.59} $ & & $67.23_{1.24}$ & $62.71_{1.16}$ & $93.32_{0.20}$ & $66.59_{1.24}$ \\
        PatchGCN \cite{Patch-GCN} & $87.13_{1.45}$ & $\mathbf{66.99_{4.42}}$ & $\underline{94.77_{0.68}}$ & $86.46_{1.80}$ & & $\underline{67.94_{1.04}}$ & $\underline{63.50_{1.06}}$ & $\underline{93.45_{0.19}}$ & $\underline{67.44_{0.93}}$ \\
        Hyper-AdaC \cite{Hyper-AdaC} & $\mathbf{87.67_{1.03}}$ & $63.27_{1.93}$ & $94.45_{0.65}$ & $86.35_{0.93}$ & & $60.98_{0.96}$ & $56.39_{0.68}$ & $91.95_{0.18}$ & $60.34_{0.91}$ \\
        bHGFN \cite{bHGFN} & $81.50_{1.80}$ & $53.94_{2.09}$ & $90.94_{0.92}$ & $78.66_{1.68}$ & & $62.08_{0.95}$ & $57.81_{0.97}$ & $92.15_{0.19}$ & $61.51_{0.99}$ \\
        \midrule
        DyHG(Ours) & $87.42_{1.24}$ & $\underline{66.82_{2.26}}$ & $\mathbf{95.08_{0.29}}$ & $\mathbf{86.59_{0.84}}$ & & $\mathbf{68.00_{1.07}}$ & $\mathbf{64.37_{0.72}}$ & $\mathbf{93.52_{0.18}}$ & $\mathbf{67.84_{0.99}}$ \\
        \bottomrule
    \end{tabular}
\end{table*}

\subsection{Case study: attention heatmap of DyHG}

\begin{figure*}[t]
\centerline{\includegraphics[width=1\linewidth, height=0.5\linewidth]{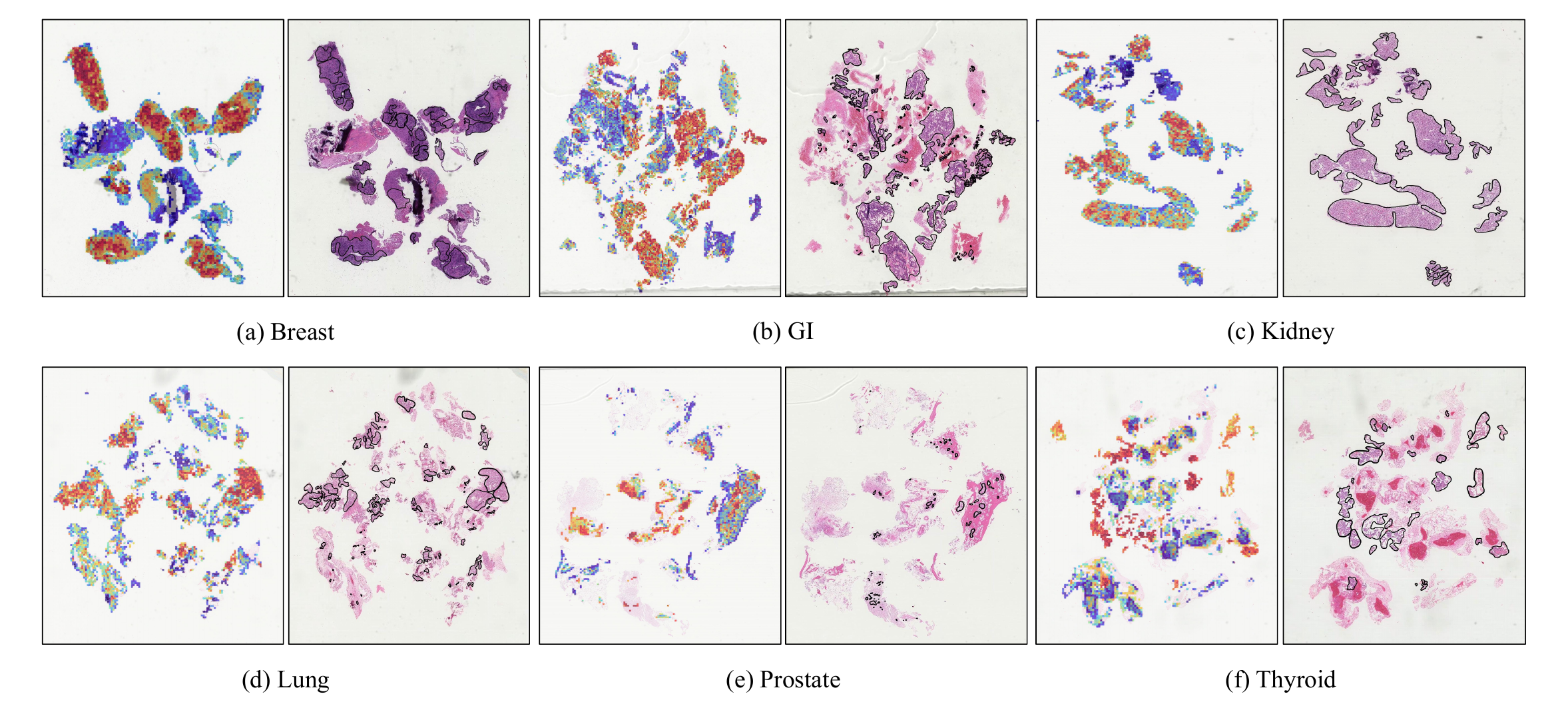}}
\caption{Visualization of the attention heatmap of DyHG (left) and the ROIs annotated by pathologists (right) for six specific tumor types, including breast (a), gastrointestinal cancer (b), kidney (c), lung (d), prostate (e), and thyroid (f).}
\label{heatmap}
\end{figure*}

To evaluate the interpretability of DyHG, we visualize the attention heatmaps generated by DyHG for six WSIs representing different primary bone cancer origins based on the attention matrix computed during the global attention pooling stage. These heatmaps are compared with the ROIs of the same WSIs annotated by pathologists, as shown in Fig.~\ref{heatmap}. It is evident that the focus areas identified by DyHG (indicated by darker regions in the heatmaps) largely align with the ROIs annotated by pathologists. This alignment demonstrates that DyHG’s decision-making process is grounded in biologically relevant regions, indicating strong interpretability.

However, it is important to note that the ROIs annotated by pathologists indicate regions where lesions are present, but do not necessarily label all subregions as being lesions. Consequently, it is reasonable for DyHG to assign lower attention (indicated by lighter colors in the heatmap) to certain parts of the annotated ROIs. For example, in the WSI belonging to the category "gastrointestinal cancer" (GI), the pathologist annotated most of the regions as ROIs. However, DyHG assigns less attention to certain areas. This may be because these regions are less distinct compared to other ROIs, suggesting that DyHG is capable of discriminating between highly representative regions of the lesion and less critical areas. This observation further highlights the strong discrimination capability of DyHG.

\begin{table}[htbp]
\centering
\caption{Dataset Statistics}
\label{d_s}
\begin{tabular}{lcc}
\toprule
\textbf{Metric} & \textbf{CAMELYON+} & \textbf{PANDA} \\
\midrule
Total WSIs & 1349 & 10210 \\
Number of Classes & 4 & 6 \\
Min Patches per WSI & 82 & 10 \\
Max Patches per WSI & 36671 & 183 \\
Avg. Patches per WSI & 5923 & 36 \\
\bottomrule
\end{tabular}
\end{table}

\subsection{Further Experiments}
To further validate the generalizability of our proposed model, we conducted comparative experiments between DyHG and several baseline methods on two publicly available datasets: CAMELYON+ \cite{camelyon+} and PANDA \cite{panda}. The CAMELYON+ dataset integrates and refines data and annotations of CAMELYON16 and CAMELYON17, providing an objective benchmark for evaluating model performance. In contrast, the PANDA dataset focuses on the Gleason grading of prostate cancer, characterized by a significantly smaller number of patches per WSI. This dataset is particularly suitable for assessing the model's effectiveness in handling WSIs with limited patch quantities. The statistical information of the two datasets is summarized in TABLE~\ref{d_s}.

We conducted experiments in identical settings, and the comparison results are presented in TABLE~\ref{tab1}. Based on the results, we derive the following observations:

(1) DyHG exhibits competitive performance in both public datasets, demonstrating its strong generalizability. Specifically, on the CAMELYON+ dataset, DyHG achieves the highest weighted F1 (86.59\%) and specificity (95.08\%) among all models, outperforming hypergraph-based baselines like Hyper-AdaC and bHGFN. Similarly, on the PANDA dataset, DyHG achieves the best results in all four metrics, including accuracy (68.00\%) and balanced accuracy (64.37\%), which highlights its ability to generalize across datasets with varying patch-level characteristics.  

(2) Graph representation learninng baselines outperform other baselines on both datasets, underscoring their ability to capture the complex biological relationships in pathological images. For instance, PatchGCN achieves competitive performance in PANDA with a balanced accuracy of 63.50\%, which is significantly higher than nongraph-based methods like ABMIL (58.52\%) and TransMIL (53.21\%). This suggests that graph structures are more effective in modeling patch-level interactions. Additionally, DyHG surpasses PatchGCN in weighted F1 (67.84\% vs. 67.44\%), demonstrating the advantage of hypergraph-based learning for finer-grained patch-level distinctions.  

(3) Notably, DyHG achieves the highest specificity on CAMELYON+ (95.08\%) and PANDA (93.52\%), which suggests its effectiveness in minimizing false positives. This can be attributed to the hypergraph structure's ability to model subtle relationships across patches, crucial for identifying small-scale lesions in WSIs. Other baselines, such as PatchGCN, also perform well in specificity, but they fall short in balanced accuracy, likely due to their limited ability to handle imbalanced patch distributions.  

(4) In terms of balanced accuracy, DyHG consistently outperforms other hypergraph models, including Hyper-AdaC and bHGFN. This improvement is particularly pronounced in PANDA, where patch numbers are smaller. The superiority of DyHG can be linked to its ability to dynamically adapt to the data distribution and refine patch-level representations during training, while other baselines rely more on static or less flexible representations, which may underperform in handling subtle variations in Gleason grades.



\section{Conclusion}
In this paper, we propose a novel dynamic hypergraph representation, DyHG, for the analysis of bone metastasis cancer. DyHG dynamically constructs hypergraphs using a low-rank strategy and Gumbel-Softmax-based sampling, enabling it to capture high-order relationships and patch-level heterogeneity in whole-slide images (WSIs). Additionally, we employ a hypergraph convolutional network that integrates node aggregation and hyperedge aggregation to facilitate information propagation and feature learning. Extensive experiments, ablation studies, and visualization results demonstrate the effectiveness, robustness, and interpretability of DyHG. In particular, the ablation study highlights the critical role of dynamic hypergraph construction and Gumbel-Softmax-based sampling in achieving superior performance, while the attention heatmap analysis confirms the biological relevance of the model's focus areas in WSIs.

In future work, we plan to explore better methods for dynamically learning hypergraph structures to further optimize the hypergraph structure. Furthermore, we plan to investigate the effectiveness of different hypergraph convolutional layers and hypergraph pooling strategies, as well as their impact on improving the scalability and adaptability of DyHG for broader pathological image analysis challenges. 

\bibliographystyle{IEEEtran}
\bibliography{main}

\end{document}